%% file: main-article.tex
\documentclass{nature_meth}
\usepackage{amssymb,amsfonts,amsmath}
\usepackage{graphicx} 
\usepackage{gensymb}

 \usepackage{multirow}
\usepackage{epsfig}
\usepackage[figoff]{figcaps}
\usepackage{hyperref}
\hypersetup{
    colorlinks=true,
    linkcolor=blue,
    filecolor=magenta,
    urlcolor=cyan,
}

\input{macros}

\usepackage{caption,setspace}
\renewcommand{\spacing}[1]{\renewcommand{\baselinestretch}{#1}\large\normalsize}
\spacing{2}

\usepackage{url}

\usepackage{graphicx}
\makeatletter
\let\saved@includegraphics\includegraphics
\AtBeginDocument{\let\includegraphics\saved@includegraphics}
\makeatother

\title{Bidirectional Generation of Structure and Properties Through a Single Molecular Foundation Model}

\author{Jinho Chang$^{1}$ and Jong Chul Ye$^{1, \dagger}$
}
\begin{document}
\setstretch{1.2}

\maketitle

\begin{affiliations}
\item Graduate School of AI, KAIST, Daejeon, South Korea
\item[] $^{\dagger}$Correspondence should be addressed to J.C.Y. (jong.ye@kaist.ac.kr)
\end{affiliations}
 
%\newpage
\vspace{-0.5cm}
\section*{Abstract}
\begin{abstract}
The recent success of large foundation models in artificial intelligence has prompted the emergence of chemical pre-trained models. Despite the growing interest in large molecular pre-trained models that provide informative representations for downstream tasks, attempts for multimodal pre-training approaches on the molecule domain were limited. To address this, we present a novel multimodal molecular pre-trained model that incorporates the modalities of structure and biochemical properties, drawing inspiration from recent advances in multimodal learning techniques. Our proposed model pipeline of data handling and training objectives aligns the structure/property features in a common embedding space, which enables the model to regard bidirectional information between the molecules' structure and properties. These contributions emerge synergistic knowledge, allowing us to tackle both multimodal and unimodal downstream tasks through a single model. Through extensive experiments, we demonstrate that our model shows remarkable capabilities in solving various meaningful chemical challenges, including conditional molecule generation, property prediction, molecule classification, and reaction prediction.

%Despite the impressive successes of deep learning approaches for various chemical problems such as property prediction, virtual screening, and {\em de novo} molecule design, separately designed models for specific tasks are usually required, and it is often difficult to synergistically combine these models for novel tasks. To address this, here we present a bidirectional molecular foundation model that can be used for both molecular structure and property inferences through a single model, inspired by recent multimodal pre-trained models. Thanks to the structure/property alignment in a common embedding space, experimental results confirm that our method leads to state-of-the-art performance and interpretable attention maps in both multimodal and unimodal tasks, including conditional molecule generation, property prediction, molecule classification, reaction prediction, etc. 

\end{abstract}

\clearpage
\setstretch{1.6}

\section*{Introduction}

%-Rise of Deep learning in various chemical domain
Capturing complex relations between chemical objects and their properties is the essence of numerous chemical challenges.
During the last decade, artificial intelligence has emerged as a promising tool in chemistry research for estimating many biochemical properties and interactions between molecules, polymers, and proteins, which are difficult to obtain experimentally\cite{solubility, polymer, pignet}. Various deep learning-based approaches in the chemical domain employed deep neural networks to extract desired characteristics like intrinsic properties, biochemical activities, and chemical reactions from raw molecule data\cite{xu_2022,paul2020artificial,reaction}. Especially, {\em de novo} molecule design has been extensively studied using recurrent networks\cite{denovo_rnn}, variational autoencoders\cite{denovo_vae,lcj_denovo1}, graph networks\cite{lcj_denovo2}, etc\cite{denovo_transformer,denovo_rlold,denovo_rl}. More recently, unsupervised learning approaches of learning better representations of the chemical inputs have been suggested \cite{chemberta, pretrain-drug,antibody_pretrain} to overcome the limitation of learning separate features for each task in a supervised manner. These recent approaches are on the same track as the concept of the foundation models that are trained with large datasets and are often considered as a new paradigm of deep learning\cite{foundationmodel,chem_fm}. 

%[unchanged paragraph, but moved from the Discussion section] 
Specifically, a concept of pre-training a neural network in a self-supervised manner for a better feature representation has been adapted for various chemical fields\cite{pretrain-drug,antibody_pretrain,chemberta}. N-Gram Graph\cite{n-gram} and GROVER\cite{grover} used a graph neural network and a graph transformer network, respectively, to obtain a pre-trained model from the molecular graph. ChemBERTa-2 \cite{chemberta2} trained a roBERTa model with 77 million molecules to build a molecular foundation model, by training the model to predict 200 different chemical property values.

%-vlp-multimodal learning
Meanwhile, in the computer vision field, multimodal pre-training methods like Vision-Language Pre-training (VLP)\cite{vlp_review} have achieved outstanding performance in downstream tasks that require an understanding of both image and  text. Most of the modern VLP models utilize Transformer\cite{transformer} architecture and its cross-attention mechanism to learn the correlation between different modalities\cite{uniter, oscar}. Moreover, several works introduced contrastive learning, which assimilates features with the same context and distances semantically unrelated features, to align image and language features in the common feature space\cite{clip, albef, coca}. VLP enables various tasks such as visual question answering\cite{vqa}, image-text retrieval\cite{retrieval_review}, text-driven image generation\cite{t2i_review}, image-driven text generation\cite{i2t_review}, etc., which are not possible using single modality foundation models.

%[unchanged paragraph, but moved from the Discussion section]
 Inspired by the success of multimodal learning, several recent works tried to obtain a better feature of a molecule by leveraging knowledge from different data representations. Winter \emph{et al.} trained a translation model between Simplified Molecular-Input Line-Entry System (SMILES) and International Chemical Identifier (InChI) key to get a feature vector with meaningful information that both molecular representations have in common\cite{smiles_inchi_vae}. Zhu \emph{et al.} used a self-supervised training method of BYOL\cite{byol} between different molecule representations of SMILES and molecular graphs to build a dual-view model\cite{graph-smiles}. However, these works introduced multimodality only for the enhancement of a molecule feature for unimodal tasks, not for the interplay between those different modalities. Furthermore, since SMILES, InChI, and graph representations contain almost identical information about the connection between atoms in a molecule, it is unlikely to expect new emergence properties by multimodal learning between these different molecule representations.

In this work, we are interested in the cross-modal comprehension between molecule structure and the associate properties, which facilitates solving meaningful tasks in many applications like property predictions, conditional molecule design\cite{denovo_review, inverseQSAR}, etc.
Taking a step further from multi-task learning methods\cite{multitask} which use the prepared properties as labels to extract general features\cite{chemberta2}, our approach regards a set of properties as a stand-alone modality that represents the input molecule and suggests that multimodal learning for molecules with this property modality can provide much more informative features.
Specifically, we propose a novel molecule Structure-Property Multi-Modal foundation model(SPMM) which allows various chemistry experiments {\em in silico}, which is pre-trained with a wide range of molecules' structures and a vector of its properties. By employing a Transformer architecture\cite{transformer}, the intramodal feature extraction and intermodal fusion can be done with self-attention and cross-attention mechanisms, respectively.

Our experimental results show that simultaneous learning of structural features with information from the associate properties through a single foundation model gives us a better representation that can be fine-tuned for various downstream tasks. Specifically, by treating both structure and property symmetrically, the model can perform bidirectional generation and prediction with a single pre-trained model, which was not possible before.

\begin{figure}[!t]
	\centering
 \centerline{\epsfig{figure=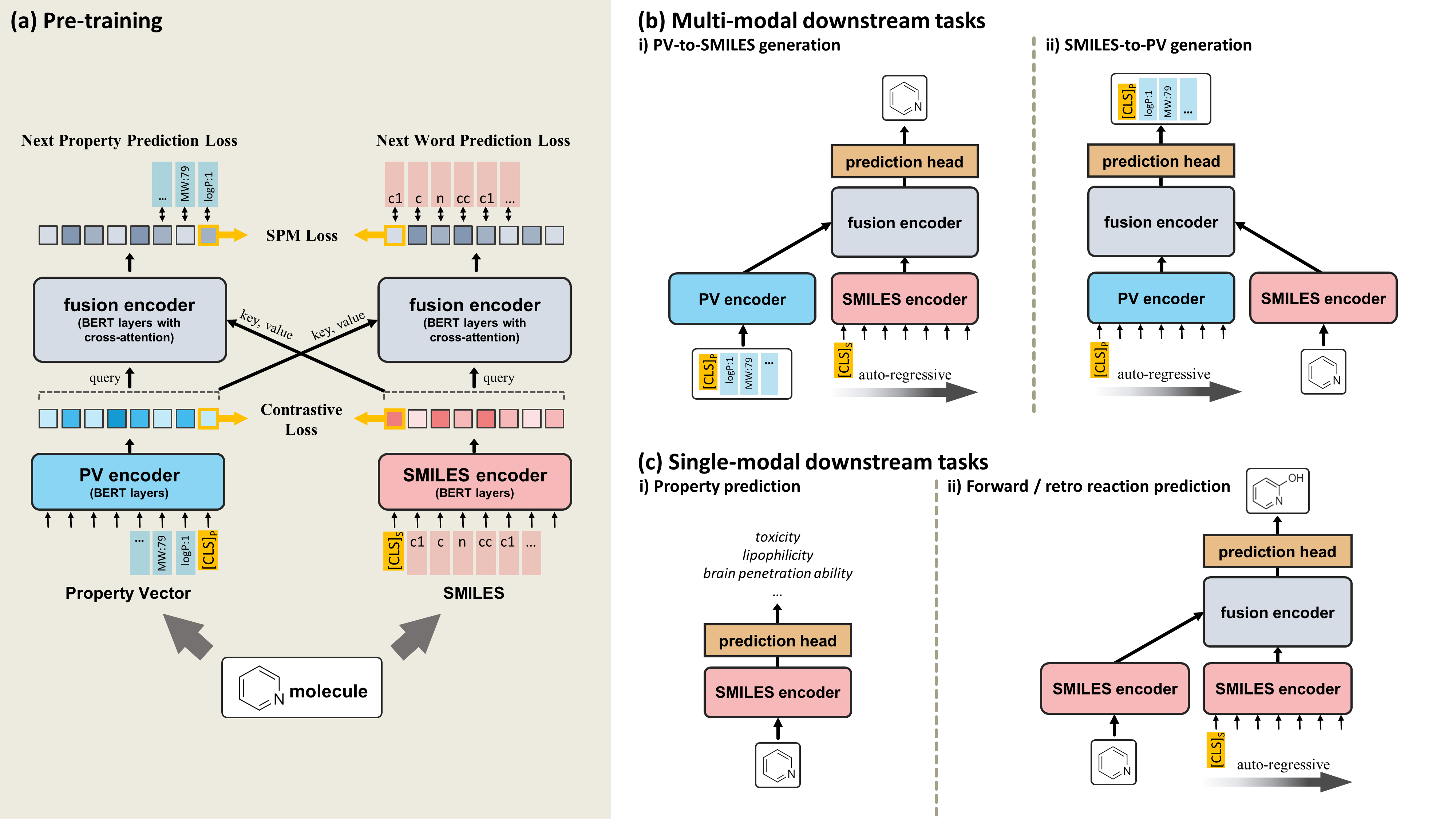, width=1\linewidth}}
	\caption{\bf\footnotesize 
(a) Overview of the model architecture and pre-training objectives of SPMM. {The contrastive loss aligns the output feature of two unimodal encoders into the same embedding space. The fusion encoder learns the relations between two modalities, trained with Next Word Prediction (NWP), Next Property Prediction (NPP), and SMILES-Property Matching loss (SPM). (b) Downstream tasks that require multimodal comprehension: i) PV-to-SMILES generation, ii) SMILES-to-PV generation. (c) Downstream tasks for single modality inputs: i) property prediction, ii) forward and retro reaction prediction.}}
	\label{fig1}
\end{figure}

Fig.~\ref{fig1}(a) illustrates the overall model architecture and training objectives for SPMM. The framework of SPMM extends the structure of the dual-stream VLP models\cite{coca,albef,xvl}. Dual-stream VLP models encode the input for each modality with a unimodal encoder, then use another encoder module to perform cross-attention by using one modality feature as a query and the other modality feature as a key/value. 
When a training molecule is given, SPMM takes the molecule's SMILES string and its property vector (PV) as multimodal data inputs as shown in Fig.~\ref{fig1}(a). The SMILES and PV are passed through their corresponding unimodal encoders, which perform self-attention where embedded inputs become the key, query, and value. 
After two unimodal features are obtained, contrastive learning aligns the SMILES and PV features into the same embedding space by assimilating the features that contain the same context. This is known to improve the model performance by making cross-modal encoding easier and guiding the unimodal encoded features to reflect more semantics of the input\cite{albef}. 
Then, the encoded SMILES and PV features are passed through the fusion encoders, which perform cross-attention between SMILES and PV features. This single fusion encoder can perform cross-attention with an alternation of its query and key/value input because the contrastive learning aligns the output of the SMILES encoder and the PV encoder into the same feature space.\cite{xvl} 
The fusion encoder is pre-trained with Next Word Prediction (NWP) for SMILES, Next Property Prediction (NPP), and SMILES-PV Matching loss (SPM). Prediction of the next component from the given transformer input is a commonly used self-supervised learning objective, and our NWP and NPP tasks make the model learn the contextual relationship between SMILES tokens and properties with the aid of the other modality’s semantic feature. Additionally, SPM predicts whether a given pair of SMILES and PV represents the same molecule or not.

%-We proposed...
Once trained, SPMM can be used for various bidirectional downstream tasks that require an understanding of both SMILES and properties like property prediction (SMILES-to-properties) and property-conditioned molecule generation (properties-to-SMILES, also referred to as inverse-QSAR\cite{inverseQSAR}) as shown in Fig.~\ref{fig1}(b). Furthermore, the pre-training objectives that we've used allow the pre-trained SPMM to be applied for single-modality tasks as well, such as molecule classification and reaction predictions (see Fig.~\ref{fig1}(c)). The pre-trained SPMM showed comparable performances to state-of-the-art models in these unimodal tasks, which suggests the model's generalization ability as a foundation model.

%\subsection{Related works}

\section*{Results}
\subsection{The model learns bidirectional comprehension between SMILES and properties.}
%\subsubsection*{Conditional PV-to-SMILES generation}
Once SPMM was pre-trained, we made the model generate SMILES with given PV inputs only,
%With the same approach as property generation, the pre-trained SPMM can also generate SMILES that agree with a given property, 
which is a crucial challenge for many chemical tasks such as {\em de novo} molecule design. As one of the major approaches for drug discovery, various methods have been suggested for generating molecules with desired properties\cite{denovo_transformer,lcj_denovo1,denovo_rl,lcj_denovo2}. In the approaches presented so far, the maximum number of simultaneously controllable properties wasn't very large. Also, the length of the input property vector cannot be changed. Whenever the target properties change, the model needs to be trained again for the new wanted conditions. In contrast, the pre-trained SPMM can take 53 properties used in pre-training as input conditions and generate molecules that satisfy all of them, without separate additional training for each property combination. Moreover, for the properties that we don't want to control, we can let the model ignore those conditions by replacing them with the [UNK] token that we used in pre-training. This is very useful because controlling all 53 input properties is not a usual scenario in practice, and is also not easy since the properties are correlated and entangled (\textit{e. g.}, `5 atoms \& 30 bonds' or `2 rings \& 5 aromatic rings' is unlikely to be a valid PV input).

To demonstrate the molecule generation capability of SPMM, we prepared a number of PV-to-SMILES generation scenarios and let the pre-trained SPMM autoregressively generate SMILES using the input properties. This process of SPMM is very similar to the sequence-to-sequence translation tasks in terms of the model pipeline (see Figure \ref{suppl2}-(a) for details), from the property sentence of PV to the molecular structure sentence of SMILES. 

The validity, uniqueness, and novelty of the generated molecules are the quantitative metrics of SPMM's molecule generation, defined as follows:
\begin{gather*}
\text{validity}=\frac{\text{\#SMILES with valid syntax}}{\text{\#generated SMILES}} \\
\text{uniqueness}=\frac{\text{\#non-duplicate valid SMILES}}{\text{\#valid SMILES}} \\
\text{novelty}=\frac{\text{\#unique SMILES not in the pre-training data}}{\text{\#unique SMILES}} 
\end{gather*}
Additionally, as a  qualitative metric to see how the generated SMILES match the property input, we measured the normalized Root Mean Square Error (normalized RMSE) between the input conditions and the generated molecules' properties. More specifically, we calculate the average of the RMSE of all controlled properties, after those values are normalized with the corresponding property's mean and standard deviation in the pre-training dataset. We note that RMSE was calculated on the normalized scale of each property because the values of the properties span multiple orders of magnitude. 

\begin{table}[!hbt]
\centering
\resizebox{\columnwidth}{!}{%
\begin{tabular}{r|r|ccc|c}
\hline
\multicolumn{1}{c|}{sampling} & \multicolumn{1}{c|}{input PV}                     & validity & uniqueness & novelty & normalized RMSE\\ \hline
deterministic                & 1,000 unseen PubChem SMILES' PV                                & 0.982$\pm$0.003    & 0.830$\pm$0.077      & 0.954$\pm$0.004  & 0.194$\pm$0.006 \\ \hline
\multirow{4}{*}{stochastic}  & full PV of the molecule \textbf{1}          & 0.882$\pm$0.004         & 0.999$\pm$0.001           & 1.000$\pm$0.000        & 0.189$\pm$0.007 \\
                             & Molecular weight=150                           & 0.913$\pm$0.008    & 0.999$\pm$0.000      & 0.932$\pm$0.005  & 0.262$\pm$0.004  \\
                             & \#ring=2, \#aromatic ring=1, TPSA=30, QED=0.8  & 0.933$\pm$0.007    & 0.999$\pm$0.001      & 0.979$\pm$0.005  & 0.343$\pm$0.009 \\
                             & no property control                            & 0.751$\pm$0.008    & 1.000$\pm$0.000      & 0.998$\pm$0.002  & -      \\ \hline
\end{tabular}
}
\vspace*{-0.5cm}
\caption{\bf\footnotesize 
Quantitative and qualitative results on various scenarios of PV-to-SMILES generation tasks, with the mean value and standard deviations. For deterministic sampling, we ran the experiment with four different random sets of 1,000 unseen PVs. In the case of stochastic scenarios, four different random seeds were used for each experiment.
}
\label{tablead1}
\end{table}

For the first PV-to-SMILES generation scenario, we prepared 1,000 PVs of SMILES from PubChem\cite{kim2021pubchem} that are not contained in the pre-training dataset and fed them to the pre-trained SPMM to generate appropriate SMILES. Here, the sampling process was done in a deterministic manner (greedy sampling): starting from the SMILES [CLS] token ([CLS]$_S$), the model predicts the probability distribution of the next token and chooses the option with the highest probability. The first row of Table ~\ref{tablead1} shows its results. Among the output of deterministic PV-to-SMILES generation for 1,000 PVs, 98.2\% of the generated output were valid SMILES. The mean RMSE of the 53 normalized properties was 0.194, which implies that the properties of the generated samples agree with the property input.

\begin{figure}[!hbt]
	\centering
 \centerline{\epsfig{figure=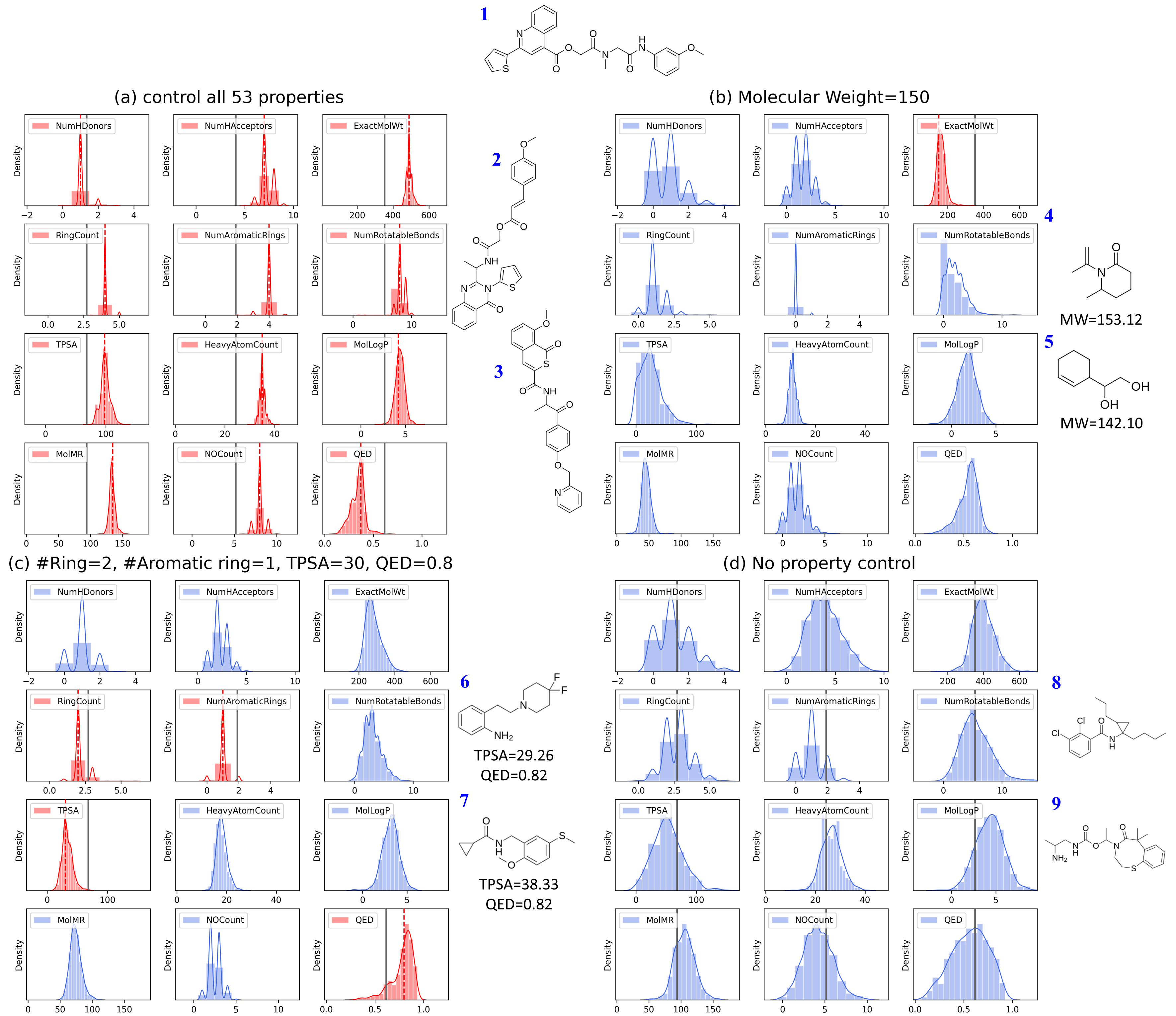, width=1\linewidth}}
	\caption{\bf\footnotesize 
Property distribution of the generated molecules with different PV inputs and [UNK] token masking. The red vertical dotted lines are the input property values, and the grey vertical lines are the mean of that property in the pre-training dataset. The controlled properties are colored in red, and uncontrolled properties (=masked with [UNK] token) are colored in blue. Due to the lack of space, only 12 out of 53 properties are shown for each case. For each PV-to-SMILES scenario, we included the structure of two of the generated molecules. (a) All 53 properties are controlled, without using the [UNK] token. The input PV was obtained from the molecule \textbf{1}. (b) Molecular Weight to 150, and the other property inputs are masked. (c) \#ring, \#aromatic ring, TPSA, and QED are controlled to 2, 1, 30, and 0.8. The other property inputs are masked. (d) Every property is replaced with [UNK] token.
}
	\label{fig4}
\end{figure}

Application fields like drug discovery often require generating multiple molecules for a single wanted target property condition. This can be done by sampling the next token stochastically from the modeled probability distribution instead of using a token with the highest probability. To verify our model's ability to generate multiple molecules from a single PV input, we generated 1,000 SMILES with stochastic sampling on a fixed PV.
Figure ~\ref{fig4} shows the property distributions of 1,000 molecules generated from a single PV input. The mode of each property distribution lands on the input property value (Fig.~\ref{fig4}-(a)). In the situation when only some of the properties are given, the model only regards the known properties while the other masked properties are not restricted (Fig.~\ref{fig4}-(b), Fig.~\ref{fig4}-(c)). SPMM can generate molecules even with no property information at all; when all input properties are replaced with [UNK] token (Fig.~\ref{fig4}-(d)), the model performs an unconditional molecule generation, and the output follows the distribution of the pre-training dataset. The validity, uniqueness, and novelty of the generated molecules under conditions in Figure ~\ref{fig4} are listed in the ``stochastic" rows of Table ~\ref{tablead1}. The validity fluctuated depending on how feasible or difficult the property input is, and it was between 0.75 and 0.9 in most cases. The uniqueness, the ratio between the number of unique molecules against the number of validly generated molecules, was almost 100\% in every condition we have experimented with. More examples of the generated molecule can be found in Supplementary result \ref{suppl4}.

\begin{figure}[!t]
	\centering
 \centerline{\epsfig{figure=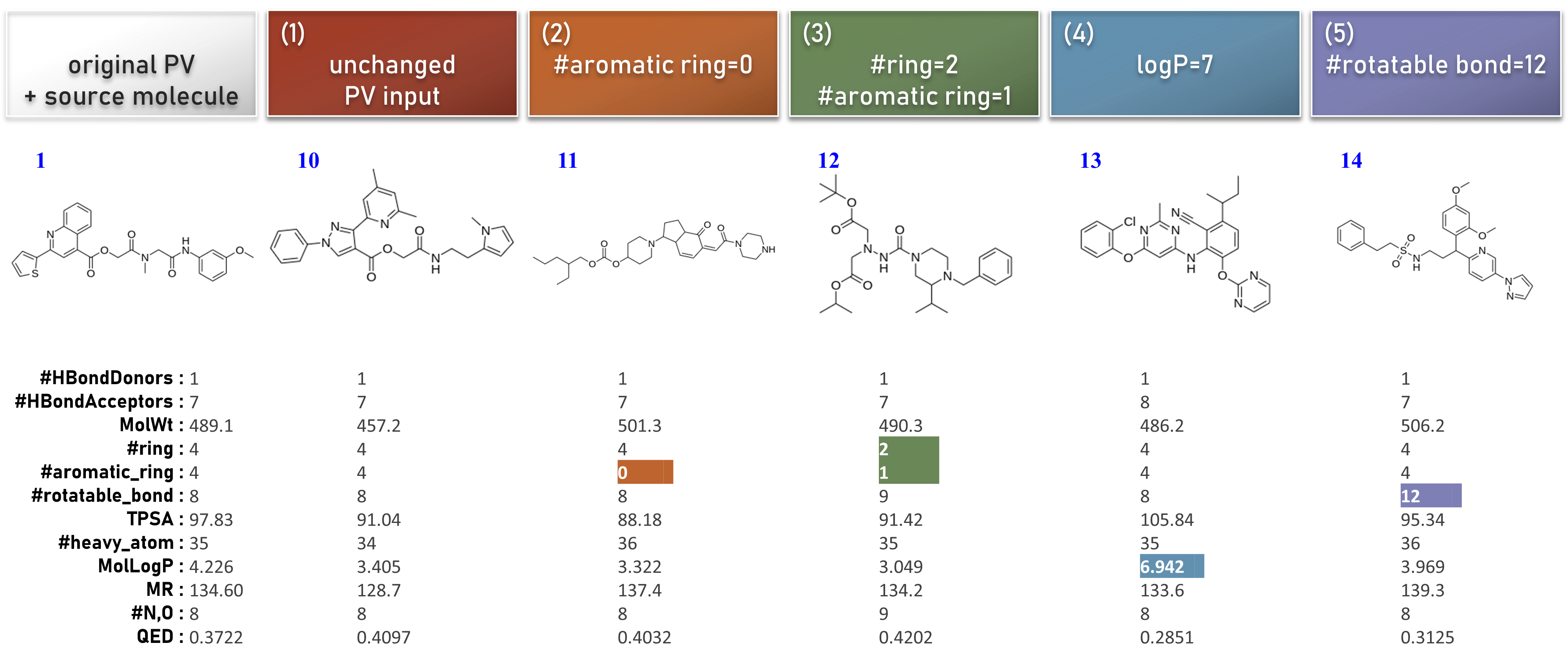, width=1\linewidth}}
	\caption{\bf\footnotesize 
Examples of molecule editing, by changing specific values from the original PV and performing PV-to-SMILES generation with it. The colored output values correspond to the changed properties from the original PV. (1) The output of the same PV of the source molecule. (2) The output when \emph{\#aromatic\_ring} is changed to 0. (3) The output when \emph{\#ring} is changed to 2 and \emph{\#aromatic\_ring} is changed to 1. (4) The output when \emph{logP} is changed to 7. (5) The output when \emph{\#rotatable\_bond} is changed to 12. For the  generation, the other 41 property conditions are masked by the [UNK] token.
}
	\label{fig3}
\end{figure}

%[unchanged paragraph, but moved down]
The aforementioned results demonstrate that SPMM can perform molecule generation with arbitrary PV inputs, which enables simple molecule designing and editing. Figure ~\ref{fig3} contains the output of the SPMM's stochastic molecule generation for five PV inputs, which all originated from the PV of the molecule \textbf{1} but four of them had certain values changed. The generated molecules follow the input modification while maintaining unmodified properties similarly. SPMM is even able to generate molecules with the out-of-domain conditions such as \emph{$\log P=7$} (note that $\sim5\%$ of the pre-training dataset has $\log P>$7).

Regarding the overall molecule generation performance of SPMM, we want to emphasize that SPMM can generate suitable SMILES for many property conditions that the model has not seen in its pre-training. When we trained SPMM without 50\% of random property masking with [UNK] token, the model only worked when all 53 properties are given since the model has not seen the partially-given properties. However, even with the technique of [UNK] token masking, the model cannot face most of the $2^{53}$ possible property combination during the pre-training process. The SPMM's ability to handle arbitrary property conditions for SMILES generation comes from treating PV as a `language with 53 words' and focusing on each property separately, not simply considering the entire property input as a single condition. This innovative approach for conditional molecule generation has never been demonstrated with the existing methods and thus can be used for many important chemical fields. 

%\subsubsection*{SMILES-to-PV generation}
\begin{figure}[!t]
	\centering
 \centerline{\epsfig{figure=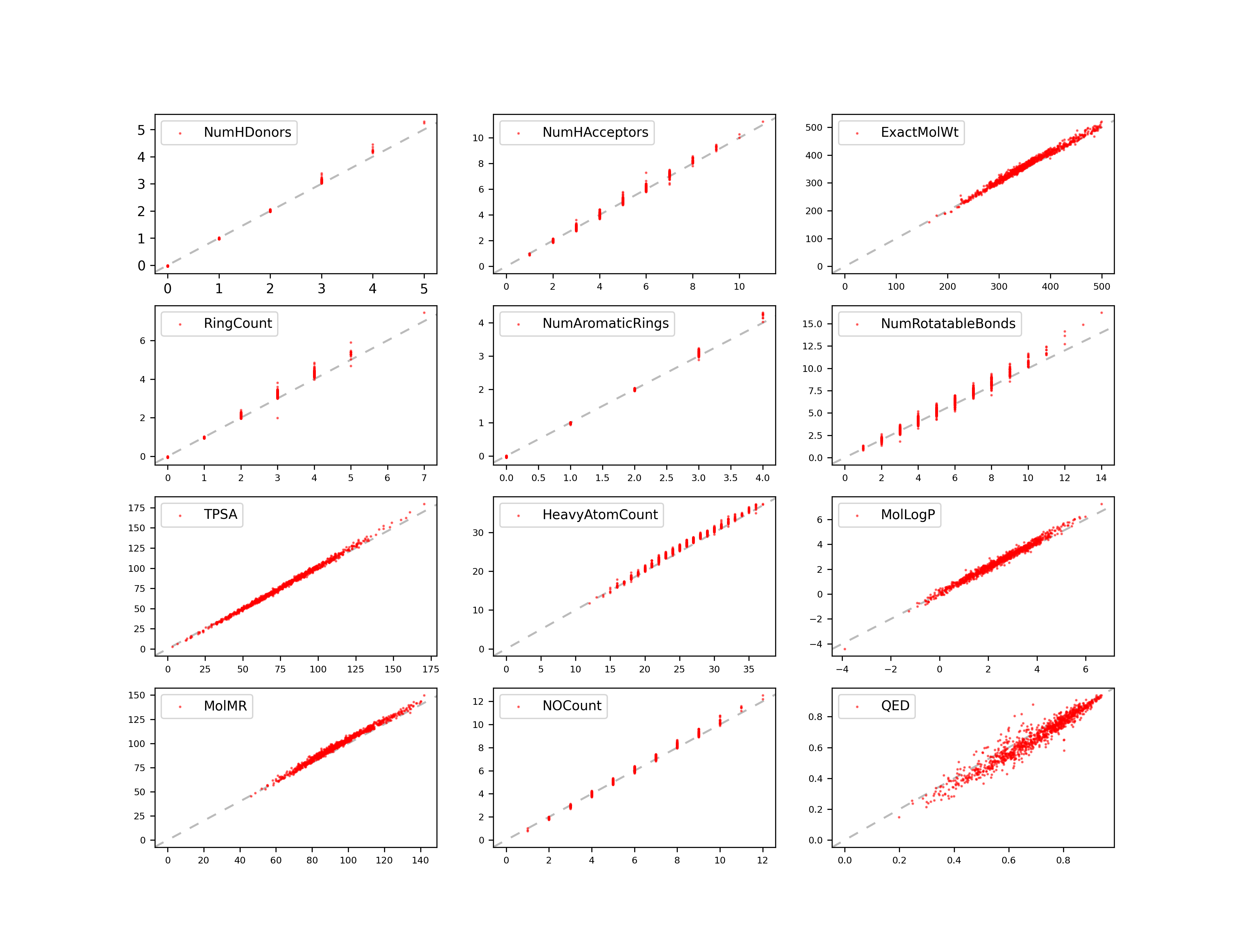, width=1.0\linewidth}}
 \vspace*{-0.5cm}
	\caption{\bf\footnotesize 
Scatter plots of the 1,000 ZINC15 molecules' real property value against the generated output, for 12 selected properties. The $x$-axis is the real property value, and the $y$-axis is the model output. The grey dotted line is the $y=x$ line.}
	\label{fig2}
\end{figure}
With the same approach as SMILES generation, the pre-trained SPMM can also be used to generate a PV with SMILES input only. 
%After the pre-training process, we made our pre-trained SPMM generate a PV with SMILES input only. 
This task is equivalent to performing 53 property predictions of a given SMILES at once. 
%No additional training is required since the pre-training with NPP already trained the model to perform property generation. 
Similar to the PV-to-SMILES generation, properties are predicted in an autoregressive manner: the model predicts the first property value using only the property [CLS] token ([CLS]$_P$), then takes all previous outputs again to get the next prediction value, and so on (see Figure \ref{suppl2}-(b)). Although 53 properties that we've used can be calculated using the Python module, the purpose of this experiment is to verify that the data-driven way of property estimation coincides with the analytic approach.

Specifically, we fed 1,000 SMILES from the ZINC15 dataset\cite{zinc15}, which are not contained in the pre-training dataset, to the pre-trained SPMM and generated their corresponding PV. Figure ~\ref{fig2} is the scatter plot of the real property value against the generated output for 12 selected properties out of 53 that we used for pre-training. It is clear that SPMM's predicted property is very close to the actual value, and most of the data point lies on the $y=x$ line. Although the model virtually has never seen a full-filled PV in the pre-training due to the 50\% of random property masking, the model could autoregressively predict all 53 properties as a whole. The mean $r^2$ score of the 53 properties was 0.932. The mean of the normalized RMSE for 53 properties was 0.118. The full scatter plot for all 53 properties with each $r^2$ score and raw RMSE is in the Supplementary Figure \ref{supp1}.

To provide an interpretation of the pre-trained SPMM's performance presented so far, we further analyzed the learned cross-modal comprehension between SMILES and property vectors by visualizing the attention scores from the pre-trained SPMM. Transformer-based models have the benefit of intuitive attention visualization that shows how the model considers the relation between the input queries and keys, by providing cross-attention scores between them. 

\begin{figure}[!hbt]
	\centering
 \centerline{\epsfig{figure=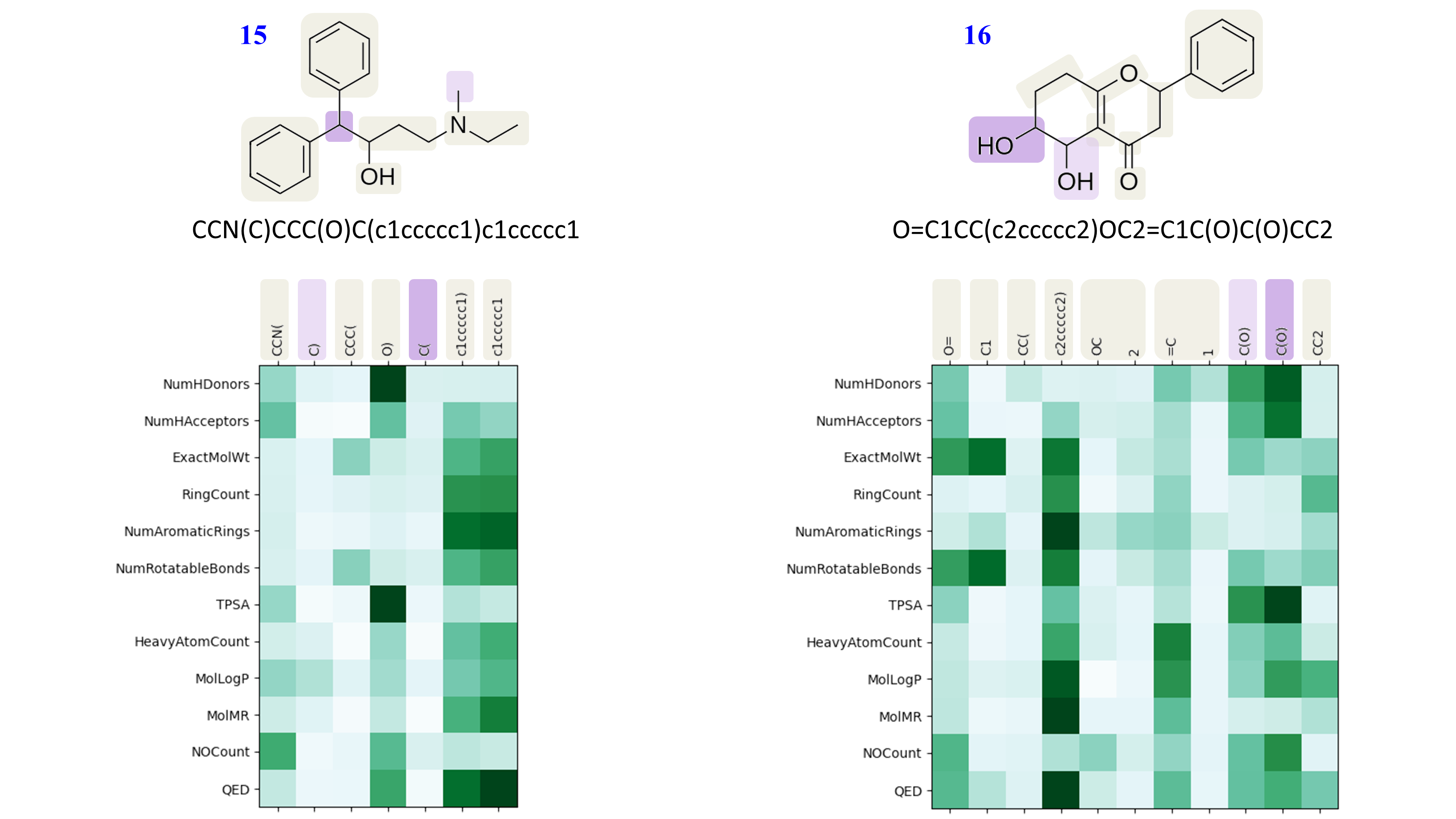, width=0.8\linewidth}}
	\caption{\bf\footnotesize 
The mean attention score from the attention heads in the SPMM fusion encoder's final cross-attention layer for two sample molecules. A darker green means a higher attention score. For the attention process, the property features were used as queries, and the SMILES features are used as keys and values. The corresponding fragments for each token are indicated with ivory boxes on the molecular structure, while fragments for duplicated tokens are color-coded with purple. We have calculated cross-attention scores for all 53 properties and SMILES tokens, but only 12 of those properties are shown.
}
	\label{av}
\end{figure}

In Figure~\ref{av}, we plotted the cross-attention score from the last fusion layer of our pre-trained SPMM when SMILES and its property vector inputs were given. Since there are multiple heads for the cross-attention, we took the mean of their attention scores. It is interesting that the aspect of cross-attention scores followed the intuitive relations between chemical properties and molecular fragments.
The properties related to hydrogen bonding (\emph{NumHDonors, NumHAcceptors}) show high attention scores for tokens with oxygen and nitrogen atoms. 
The property \emph{RingCount} focuses on the tokens that are involved with rings while showing weak attention to side groups, and the property \emph{NumAromaticRings} only gives high attention score to the components of aromatic rings. 
When different SMILES tokens played a similar role in the molecule such as `\emph{c1ccccc1)}' and `\emph{c1ccccc1}' in the molecule \textbf{15}, their attention patterns were similar as well. This result demonstrated that SPMM could capture the relations between molecule structures and chemical properties without explicitly-given supervision between them.

\subsection{Generalization ability as a molecular foundation model.}
So far, we have demonstrated that the pre-trained SPMM can be applied to tasks that require an understanding of the relationship between SMILES and properties. However, we can also employ the pre-trained SPMM for challenges that only use SMILES data, such as molecular property prediction. One advantage of having a dual-stream VLP model structure is that the SPMM's multimodal pre-training process includes adjusting the output of one unimodal encoder to contain contextual information from the other modality, by aligning it with the other unimodal encoder's output. This implies that the SMILES encoder output is a unimodal representation vector, that not only embeds the input molecule's structural information but it's also enhanced by its property information. 

We have analyzed if our pre-trained model had learned an informative representation that can be readily used for other tasks, even for a single modality. So we only utilized the SMILES encoder of pre-trained SPMM (see Supplementary Figure ~\ref{suppl2}-(c)) and made a benchmark study on nine MoleculeNet\cite{moleculenet} downstream tasks and a Drug-Induced Liver Injury (DILI) prediction task. Each MoleculeNet task is a regression or classification task for pharmaceutical/biochemical applications like solubility, toxicity, and brain penetrability. 
The DILI classification task was done to overcome the potential limitation of open databases\cite{nonadditivity1, nonadditivity2} and verify if SPMM could be extended to more complex endpoints. The task is to classify whether the given molecule has a risk of causing liver injury. Since many proposed DILI machine learning models have built their dataset rather than using common benchmarks, we took the dataset preparations from a known publication\cite{dili2018} and compared the performance with it for a fair evaluation.

Table ~\ref{table2} contains the performance of SPMM and other models for MoleculeNet. Using only 6 BERT encoder layers, SPMM showed comparable performances with state-of-the-art models for all tasks. It achieved the best performance for Clearance, BBBP, and Clintox tasks, showing its capability as a foundation model. We've also observed that the score of our model dramatically decreased without pre-training. SPMM also outperformed the proposed 5-ensemble models on the DILI classification task under the same data preparation, which was not the case for the naive BERT layers without SPMM pre-training.

\begin{table}[!hbt]
\centering
\resizebox{\columnwidth}{!}{%
\begin{tabular}{r|ccccc|lccc}
\hline
                  & \multicolumn{5}{c|}{regression{[}RMSE, $\downarrow${]}}                                                                                                              & \multicolumn{4}{c}{classification{[}AUROC in $\%$, $\uparrow${]}}                               \\ \cline{2-10} 
Dataset           & \multicolumn{1}{l}{Delaney ESOL} & \multicolumn{1}{l}{LIPO} & \multicolumn{1}{l}{Freesolv} & \multicolumn{1}{l}{BACE} & \multicolumn{1}{l|}{Clearance} & BBBP                     & \multicolumn{1}{l}{BACE} & \multicolumn{1}{l}{Clintox} & \multicolumn{1}{l}{SIDER}\\
\#data            & \multicolumn{1}{l}{1128}         & \multicolumn{1}{l}{4200} & \multicolumn{1}{l}{642}      & \multicolumn{1}{l}{1513} & \multicolumn{1}{l|}{837}          & 2039                     & \multicolumn{1}{l}{1513} & \multicolumn{1}{l}{1478} & \multicolumn{1}{l}{1427}    \\
\#task            & \multicolumn{1}{l}{1}            & \multicolumn{1}{l}{1}    & \multicolumn{1}{l}{1}        & \multicolumn{1}{l}{1}    & \multicolumn{1}{l|}{1}         & 1                        & \multicolumn{1}{l}{1}    & \multicolumn{1}{l}{2} & \multicolumn{1}{l}{27}      \\ \hline
D-MPNN\cite{d-mpnn}            & 1.050$\pm$0.008      & \underline{0.683}$\pm$0.016     & 2.082$\pm$0.082                        & 2.253$^*$                         & 49.754$^*$                               & \multicolumn{1}{c}{71.0$\pm$0.3} & 80.9$\pm$0.6      & 90.6$\pm$0.6  &57.0$\pm$0.7           \\
N-GramRF\cite{n-gram}            & 1.074$\pm$0.107       & 0.812$\pm$0.028     & 2.688$\pm$0.085                        &1.318$^*$       & 52.077$^*$        & \multicolumn{1}{c}{69.7$\pm$0.6} & 77.9$\pm$1.5                     & 77.5$\pm$4.0  &\underline{66.8}$\pm$0.7           \\
N-GramXGB\cite{n-gram}            & 1.083$\pm$0.082    & 2.072$\pm$0.030   & 5.061$\pm$0.744              &-          &-            & \multicolumn{1}{c}{69.1$\pm$0.8} & 79.1$\pm$1.3       & 87.5$\pm$2.7   &65.5$\pm$0.7                     \\

PretrainGNN\cite{pretraingnn}            & 1.100$\pm$0.006         & 0.739$\pm$0.003          & 2.764$\pm$0.002      &-       &-           & \multicolumn{1}{c}{68.7$\pm$1.3} & 84.5$\pm$0.7         & 72.6$\pm$1.5    &62.7$\pm$0.8                    \\
GROVER$_{large}$\cite{grover}            & 0.895$\pm$0.017       & 0.823$\pm$0.010       & 2.272$\pm$0.051      &-       &-          & \multicolumn{1}{c}{69.5$\pm$0.1} & 81.0$\pm$1.4          & 76.2$\pm$3.7  &65.4$\pm$0.1      \\
ChemRL-GEM\cite{chemrl_gem}            & \textbf{0.798}$\pm$0.029     & \textbf{0.660}$\pm$0.008     & \textbf{1.877}$\pm$0.094       &-       &-         & \multicolumn{1}{c}{72.4$\pm$0.4} & \underline{85.6}$\pm$1.1    & 90.1$\pm$1.3  & \textbf{67.2}$\pm$0.4            \\
ChemBERTa-2$_{\text{(MTR-77M)}}$\cite{chemberta2} & 0.889$^*$       & 0.798$^*$   &-      & 1.363$^*$       & 48.515$^*$    & \multicolumn{1}{c}{72.8$^*$} & 79.9$^*$      & 56.3$^*$   &-               \\
MolFormer$^{\dagger}$\cite{molformer}            & 0.880$\pm$0.028         & 0.700$\pm$0.012          & 2.342$\pm$0.052        & \textbf{1.047}$\pm$0.029   & \underline{43.175}$\pm$1.537        & \multicolumn{1}{c}{\underline{73.6}$\pm$0.8} & \textbf{86.3}$\pm$0.6            & \underline{91.2}$\pm$1.4      & 65.5$\pm$0.2                     \\
\hline
SPMM(w/o pre-train)      & 1.272$\pm$0.015           & 1.009$\pm$0.021          & 3.018$\pm$0.179        & 1.675$\pm$0.010             & 53.544$\pm$0.312           & \multicolumn{1}{c}{66.6$\pm$0.3} & 78.7$\pm$2.6      & 76.3$\pm$1.5   &57.1$\pm$1.6                     \\
%SPMM(10m)             & \underline{0.810}                            & 0.706                    & \textbf{1.859}                        & \textbf{1.108}                    & \textbf{44.752}                         & \multicolumn{1}{c}{\textbf{73.3}} & 83.0                     & \textbf{91.0}  & 64.7                      \\ \hline 
SPMM         &  \underline{0.818}$\pm$0.008     & 0.692$\pm$0.008     & \underline{1.907}$\pm$0.058       & \underline{1.096}$\pm$0.011        & \textbf{42.841}$\pm$1.251             & \multicolumn{1}{c}{\textbf{74.9}$\pm$0.8} & 84.8$\pm$0.3      & \textbf{91.8}$\pm$0.9       & 65.5$\pm$0.9        \\ \hline 
\end{tabular}%
}
\vspace*{-0.5cm}
\caption{\bf\footnotesize 
Benchmark results on MoleculeNet downstream tasks. The best performance for each task was written in bold, and the second-best performance was underlined. For each task, we fine-tuned our model in four random seeds and recorded the mean and the standard deviation of those results. The benchmark model results were taken from ChemRL-GEM and ChemBERTa-2. $^*$The standard deviation cannot be found in the source of the benchmark results. $^{\dagger}$Unofficial results, obtained from the official checkpoint under our data preparation.}

	\label{table2}
\end{table}

\begin{table}[!hbt]
\centering
\resizebox{\columnwidth}{!}{%
\begin{tabular}{r|cccc}
\hline
model                           & Acc in \%[$\uparrow$]   & Selectivity in \%[$\uparrow$] & Specificity in \%[$\uparrow$] & AUROC in \%[$\uparrow$] \\ \hline
Ai \emph{et al.}\cite{dili2018} (best single model on training set) & 81.1        & 81.0              & 81.5              & 89.6        \\
Ai \emph{et al.} (5-ensemble)& 84.3        & \textbf{86.9}              & 75.4              & 90.4        \\ \hline
%SPMM(w/o pre-train)              & 72.6$\pm$2.8        & 70.6$\pm$2.8                  & 79.2$\pm$2.9                  & 82.0$\pm$0.7            \\
%SPMM                            & \textbf{84.7}$\pm$1.1 & 85.7$\pm$1.6       & \textbf{81.6}$\pm$3.1       & \textbf{91.2}$\pm$0.3 \\ \hline
SPMM(w/o pre-train)              & 72.6       & 70.6             & 79.2        & 82.0        \\
SPMM                   & \textbf{84.7} & 85.7     & \textbf{81.6}      & \textbf{91.2} \\ \hline
\end{tabular}
}
\vspace*{-0.5cm}
\caption{\bf\footnotesize 
The DILI classification task performance of Ai \emph{et al.}\cite{dili2018} and SPMM. The best performance for each metric was written in bold. %We recorded the mean and the standard deviation of the result of four random seeds.
}
\label{table_lidi}
\end{table}

We also trained SPMM for the forward and retro-reaction prediction tasks, which require the model to predict the product SMILES from the reactant SMILES and vice versa. Regarding both tasks as sequence-to-sequence generation, the model pipeline for these reaction prediction tasks is the same as the PV-to-SMILES generation tasks, except the PV encoder is replaced with the SMILES encoder (see Supplementary Figure ~\ref{suppl2}-(d)). The detailed task definition and dataset preparation are described in the Methods section. 

\begin{table}[]
\centering
\resizebox{0.65\columnwidth}{!}{%
\begin{tabular}{r|cc|cccc}
\hline
\multicolumn{1}{l|}{forward prediction}  & \multicolumn{2}{c|}{molecule modality}  & \multicolumn{4}{c}{top-k accuracy in \%[$\uparrow$]}     \\ \cline{2-7} 
            &string-based &graph-based    & \multicolumn{1}{l}{k=1} & \multicolumn{1}{l}{k=2} & \multicolumn{1}{l}{k=3} & k=5  \\ \hline
Molecular Transformer\cite{molecular_transformer} &O & & 88.7                     & 92.1                     & 93.1                     & 94.2 \\
Augmented Transformer\cite{aug_transformer} &O &  & 90.6                     & \underline{94.4}                     &-                          & \underline{96.1} \\
Chemformer$_{large}$\cite{chemformer} &O &             & \underline{91.3}                     &-                          &-                          & 93.7 \\
Graph2SMILES\cite{graph2smiles} &O &O           & 90.3                     &-                          & 94.0                     & 94.8 \\
MEGAN\cite{megan} & &O                  & 86.3                     & 90.3                     & 92.4                     & 94.0 \\
LocalTransform\cite{reaction} & &O         & 90.8                     & \textbf{94.8}                     & \textbf{95.7}                     & \textbf{96.3} \\ \hline
SPMM &O &                   & \textbf{91.5}                     & 93.4                         & \underline{94.6}           & 95.3     \\ \hline
\end{tabular}
}

\resizebox{0.45\columnwidth}{!}{
\begin{tabular}{r|ccc}
\hline
\multicolumn{1}{l|}{retro-reaction prediction}     & \multicolumn{3}{c}{top-k accuracy in \%[$\uparrow$]}         \\ \cline{2-4} 
              & \multicolumn{1}{l}{k=1} & \multicolumn{1}{l}{k=5} & k=10  \\ \hline
SCROP\cite{scrop} & 43.7                     & 65.2                               & 68.7 \\
Two-way Transformer\cite{tw-transformer} & 47.1                     & \underline{73.1}                 & \underline{76.3} \\
Augmented Transformer\cite{aug_transformer}           & 48.3          &\textbf{73.4}           & \textbf{77.4} \\
Chemformer$_{large}$\cite{chemformer}          & \textbf{54.3}        & 62.3            & 63.0 \\ \hline
SPMM                  & \underline{53.0}        & 67.4            & 70.3     \\ \hline
\end{tabular}
}
\caption{\bf\footnotesize 
The performance of SPMM and other works on the forward and retro-reaction prediction task. For the retro-reaction prediction task, we only prepared the benchmark results of string-based models. The highest accuracy is written in bold, and the performance of the runner-up model is underlined. The benchmark model results are from the paper of LocalTransform\cite{reaction} and Chemformer\cite{chemformer}.}
\label{table3}
\end{table}

Table ~\ref{table3} shows the performances of SPMM and other benchmark models on forward and retro-reaction prediction tasks. Although the reaction prediction tasks are not the best scenario for the property-emergence features to play significant roles, SPMM showed the highest top-1 accuracy in the forward-reaction task with a relatively small pre-training data size (\emph{i.e.} 20M molecules, compared to 100M molecules of Chemformer). SPMM also achieved the second-best top-1 accuracy among the string-based retro-reaction task models.

\section*{Discussion}
In this work, we proposed a transformer-based multimodal chemical foundation model SPMM. The proposed model allows for bidirectional generation/prediction of molecular structure and properties, as well as unimodal tasks like reaction prediction.
During the process, we introduced a method of treating property collections as a language so that the model could learn the relationship between SMILES tokens and each property independently. We demonstrated that pre-trained SPMM showed remarkable performances in problems for interactions between SMILES and PV domains. And not only for multimodal challenges but even its unimodal feature for SMILES, SPMM also provides a useful representation that can be fine-tuned for many molecular downstream tasks. It is important to note that all of these results were obtained with a pre-training of 20 million molecules, which is relatively small compared to other large pre-training approaches and still has room for better performance with more data and parameters. We also note that we've gathered our 53 properties to let them cover the widest range possible, rather than paying the best effort to select the most effective combination of properties. This implies the proposed structure-property multimodal training can be flexibly adopted with different property selections, according to the given specified scenarios.

Despite the noticeable performances of SPMM, it has several chances for improvement. 
One of those comes from using the SMILES notation. Although SMILES can contain full details about the 2D structure of the molecule, the information on how atoms and bonds are connected only exists implicitly. Also, a slight modification in molecular structure can be a drastic change in SMILES. Graph format is another widely used modality for molecule representation that contains the explicit information of the adjacency matrix, which can be an alternative for SMILES.
Another limitation in our current SPMM is that the 53 properties we used happen to be invariant with the changes in the stereochemistry of the given molecule. It is known that considering stereochemistry plays a crucial part in various biochemical tasks. However, the 53 properties we used cannot provide any knowledge about stereochemical information since their values are unchanged in different stereoisomers. This makes the SMILES encoder output of different stereoisomers converge since the contrastive loss aligns them to the same PV feature. We believe this is the prominent factor that lowered the performance of SPMM in MoleculeNet tasks, which could be resolved by using more properties that reflect the molecule's stereochemistry. 
Moreover, validation through wet-lab experiments to verify the model's predicted/generated properties is another possible further study.
Overcoming these drawbacks of the current study and making the model more applicable to other chemical tasks could be the works for the future.

Nevertheless, we believe that our approach can provide a pre-trained model capable of encompassing each input domain and their multimodal domain simultaneously, which has a vast potential utility.
We expect this approach to be applied to more various and practical chemical situations by using broader and richer molecular modalities, and possibly, different biochemical domains like polymers and proteins.

\section*{Methods}

\subsection{Handling SMILES and property values as a language.}
Molecules can be represented with various formats such as fingerprints, strings like SMILES, InChI, or a molecular graph. Since these different notations contain almost the same information about complete molecular structure, % and defining multimodal tasks on these representations are relatively trivial, and 
we employed SMILES to describe a molecule structure. SMILES is a sequence of characters that represents the connection structure of the molecule. Many researchers treat SMILES as a variant of language data and utilize a concept of language models for chemical tasks on SMILES data\cite{denovo_transformer,chemberta2,deepdl}.

\begin{figure}[!t]
	\centering
 \centerline{\epsfig{figure=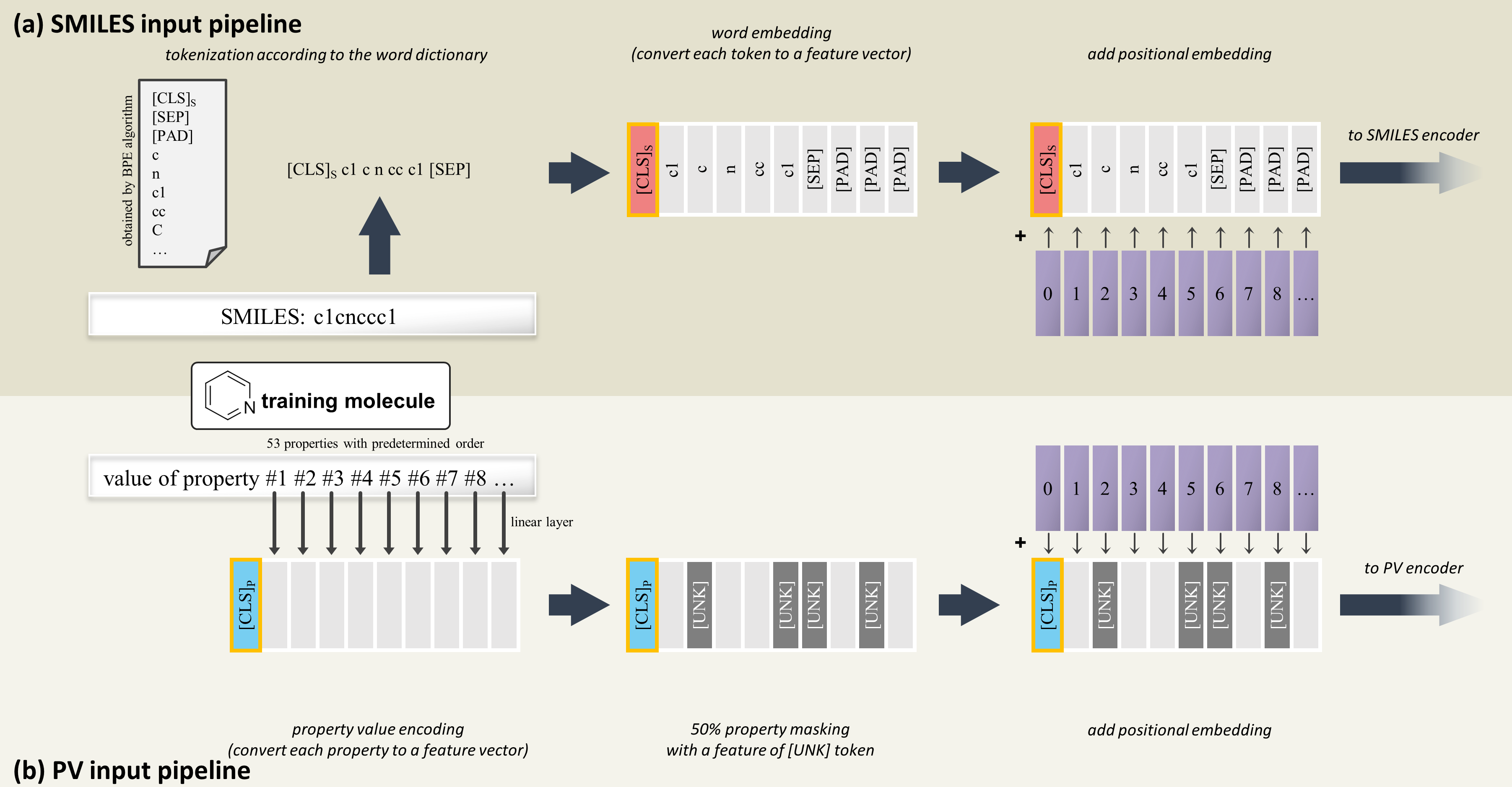, width=1\linewidth}}
	\caption{\bf\footnotesize 
Embedding process for SMILES and the corresponding PV.}
	\label{method2}
\end{figure}

Figure ~\ref{method2}-(a) illustrates our embedding procedure for the input SMILES. The raw SMILES string is tokenized by the tokenizer and embedded by the SMILES encoder with the [CLS]$_S$ token and the [SEP] token. 
Here, [CLS] token is a special token attached to the beginning of every input sequence\cite{bert}. Although the [CLS] token itself doesn’t contain any meaning, the bidirectional attention mechanism of the model allows the [CLS] token to contain contextual information of the entire input. Once the model is pre-trained, the [CLS] token output of the given sequence can be considered as an input representation vector and be used for classification/regression downstream tasks, as in many BERT variations for images\cite{dino, vit} and VLP\cite{albef}.

In the SMILES tokenization, our tokenizer tokenizes a given SMILES into fragments that are contained in a prepared token dictionary of 300 subwords. This dictionary was obtained from the pre-training data SMILES corpus by the BPE algorithm\cite{bpe}, which starts from a set of simple characters and iteratively appends the most frequent token pairs as a merged subword. Being widely adopted for various language models\cite{roberta,t5}, the BPE algorithm has provided a subword dictionary containing common functional groups and substructures like benzene rings, carbonyl groups, two-letter atoms, and amino groups. Compared to naive character-wise tokenization which considers each character as a separate token, the merged subwords help the model's chemical inference for chemical groups and reduce the total number of tokens.

Meanwhile, a set of chemical properties does not change its carrying information by changing the internal order, but they certainly have correlations between the properties. And it is known that a transformer architecture also performs well for different modalities like images, by giving specific order to its components and treating them as a sequence. For this work, we built a PV for each molecule that contains 53 molecular properties and considered this as a sentence with a length of 53. These properties from the RDKit python module\cite{rdkit} cover a wide range from simple ones, such as the number of rings and molecular weight, to complex properties like solubility, TPSA, and druggability. 

The transformer architecture of our model considers each element of PV as a token to perform the attention mechanism, which is equivalent to regarding PV as a semi-sentence of 53 properties. Although the size of the vocabulary is more limited and their order is fixed compared to natural language, it provides much more precise and compact information about the 53 properties.
One benefit of regarding PV as a language is that we do not have to collect all elements to build a valid PV. In contrast to a simple vector input, some property elements can be removed or masked in our approach.

Figure ~\ref{method2}-(b) shows our embedding procedure for the input PV. Each property element in the PV is a numerical value and normalized with the mean and standard deviation of that property. The order of these 53 properties is predetermined. Each value in the PV is encoded to a feature vector using a linear layer as a value encoding. Then we randomly replace 50\% of the property features into the [UNK] token, which is the special token utilized to simulate that the property is unknown. This is possible since there is no problem in describing a molecule using only a part of these properties. Random property feature masking prevents the model from overly dependent on the specific property, has the effect of data augmentation, and improves the model's generalization ability. Although every property we used in this work can be easily and thoroughly prepared by the computer, this might not be the case for other properties in real-world situations. SPMM still can be trained when some properties for certain training molecules are not known, by replacing those unknown properties with the [UNK] token. On top of the randomly-masked value encoding, we added a positional encoding similar to that in BERT. Since a PV explicitly contains the values only, this positional embedding provides information about what property each value corresponds to. Also, because of the pre-defined order of these properties, this position embedding is equivalent to giving a unique index for each property and adding an embedding of that corresponding index. 
Then we pass the final result to the PV encoder with the [CLS]$_P$ token.

\subsection{Pre-training objectives.}
Contrastive learning aims to learn better unimodal representation by aligning the features from different modalities into the same feature space\cite{clip}. When the encoded features of [CLS] tokens of SMILES $S$ and PV $P$ are given as $S_{cls}$ and $P_{cls}$, we calculate the similarity function $sim(S,P)$ and $sim(P,S)$ as:
\begin{align}
\label{Eq1}
sim(S,P)=(h_S(S_{cls}))^\intercal h_P(P_{cls}), \quad sim(P,S)=(h_P(P_{cls}))^\intercal h_S(S_{cls})
\end{align}
where $h_S$ and $h_P$ are the linear projection + normalization layer for SMILES and property vector, respectively.
Now, for a given pair of $S$ and $P$, we calculate the SMILES-to-PV and PV-to-SMILES intermodal similarities as follows\cite{albef, clip}:
\begin{align}
\label{Eq2}
s_{s2p}={\exp{(sim(S,P)/\tau)} \over \Sigma_{n=1}^{N}{\exp{(sim(S,P_n)/\tau})}}, \quad s_{p2s}={\exp{(sim(P,S)/\tau)} \over \Sigma_{m=1}^{M}{\exp{(sim(P,S_m)/\tau})}}
\end{align}
where $M$ and $N$ are the total numbers of SMILES and PV used in the loss calculation. Here, $\tau$ is a learnable temperature parameter, which has a sharpening effect by exaggerating the similarity difference. 
The intramodal similarities can be calculated in the same way.
\begin{align}
\label{Eq3}
s_{s2s}={\exp{(sim(S,S)/\tau)} \over \Sigma_{m=1}^{M}{\exp{(sim(S,S_m)/\tau})}}, \quad s_{p2p}={\exp{(sim(P,P)/\tau)} \over \Sigma_{n=1}^{N}{\exp{(sim(P,P_n)/\tau})}}
\end{align}

The overall contrastive loss is defined using the cross-entropy loss $H$ and one-hot similarity $y$, which contains 1 for pairs originating from the same molecule and contains 0 otherwise.
\begin{align}
\label{Eq4}
L_{contrastive}={1\over 2} (H(y_{s2p},s_{s2p})+H(y_{p2s},s_{p2s})+H(y_{s2s},s_{s2s})+H(y_{p2p},s_{p2p}))
\end{align}
Following the recent contrastive loss application in VLP\cite{moco}, we build the SMILES and PV queues that store the $k$ most recent SMILES and PV instances and use them for contrastive loss. We set our queue size $k$ to 32,768.

Next Word Prediction (NWP) trains the model to predict the $(n+1)$-th SMILES token when $0\sim n$-th tokens and the corresponding PV are given. Predicting the next token is a common objective for training language models, known for being utilized in the pre-training of GPT\cite{gpt3}. This can be done with a single flow for each SMILES by applying a causal mask in the self-attention of the SMILES encoder and the fusion encoder. Let $S=\{s_0, s_1, \dots,s_n\}$ and $P$ denote the input SMILES and the corresponding PV, and $p^{NWP}(s_n|s_{0:n-1},P)$ denote the model's predicted probability distribution of the $n$-th token with given $P$ and $0\sim(n-1)$-th SMILES tokens. The loss for NWP is defined as follows:
\begin{align}
\label{Eq5}
L_{NWP}=\sum^{n}_{i=1}H(y_n^{NWP}, p^{NWP}(s_n|s_{0:n-1},P))\ 
\end{align}
where $y_n^{NWP}$ is a one-hot label for the $n$-th SMILES token $s_n$.
% since $S$ itself becomes a label for the next word prediction task.

We applied a similar concept of NWP for the property vector as Next Property Prediction (NPP). NPP makes the model predict the next property value using its corresponding SMILES and the previous properties. Since each property element is a numerical value, we replaced the cross-entropy loss in NWP with mean-square-error loss. When $S$ and $P=\{p_0,p_1,\dots,p_n\}$ denotes the input SMILES-PV pair and $\hat{p_n}(p_{0:n-1},S)$ denotes the model's predicted next property values with causal mask in the PV and the fusion encoder, the loss for NPP is given as follows:
\begin{align}
\label{Eq6}
L_{NPP}=\sum^{n}_{i=1}(p_n-\hat{p_n}(p_{0:n-1},S))^2
\end{align}
In NPP, the model does not predict the property value if it is replaced with [UNK] token.

SMILES-Property Matching (SPM) learns if a given SMILES-PV pair $(S, P)$ is matched or not. We concatenate the feature of [CLS]$_S$ and [CLS]$_P$ token from the fusion encoder output and pass this through a linear-layer SPM head. When $p^{SPM}(S,P)$ is the output of the SPM head, the SPM loss can be defined as
\begin{align}
\label{Eq7}
L_{SPM}=H(y^{SPM}, p^{SPM}(S,P)) 
\end{align}
where $y^{SPM}$ is a one-hot vector for a binary label of SMILES-PV matching; the label is 1 if $S$ and $P$ originated from the same molecule and 0 otherwise.
%moved
To build negative samples for SPM, we randomly select a ``negative" pair for each SMILES and PV instance from the other modality and match them as negative pairs. 
This negative pair selection was done by hard-negative mining, which gives a higher chance of being selected as a negative pair for instances that has a higher similarity of Eq.~\eqref{Eq2} but isn't a positive match.
This makes the training more difficult and forces the model to learn how to distinguish similar instances.

%+EMA momentum teacher for contrastive+mlm.
In contrastive learning, using a one-hot label could be too strict since it regards all instances that came from other pairs as equally-negative instances. However, some PVs might agree with many SMILES, not only one SMILES that they're paired with. Even SMILES can be matched with different PVs since there's a 50\% of masking in a PV (\textit{e}.\textit{g}.,  ``\emph{MW}=[UNK], \emph{logP}=2.1, \emph{\#atom}=12" and ``\emph{MW}=78, \emph{logP}=2.1, \emph{\#atom}=[UNK]" both explain Benzene, even if they came from different molecules).
A similar problem also occurs for NWP. Sometimes there could be multiple sensible options for being the next token, but using a one-hot label for ground truth might ignore this.

To resolve this issue, we built the momentum teacher model\cite{moco,albef} and utilized its output for contrastive learning and NWP. The momentum teacher performs a knowledge distillation by providing a pseudo-label that reflects how the teacher model comprehends.
Specifically, the label for the contrastive learning and NWP are mixed with the momentum model's output $s_{*, momentum}(*\in\{s2p,p2s,s2s,p2p\})$ and $p^{NWP}_{momentum}(s_n|s_{0:n-1},P)$, with an adjusting hyperparameter $\alpha$. The detailed formulas for utilizing the momentum model for contrastive learning and NWP are described in Eq.~\eqref{Eq9}$\sim$\eqref{Eq10} and Eq.~\eqref{Eq11}$\sim$\eqref{Eq12}. 
\begin{align}
\Tilde{y}_{*}&=(1-\alpha)y_{*}+\alpha s_{*, momentum} \quad (*\in\{s2p,p2s,s2s,p2p\})\label{Eq9}\\
\Tilde{L}_{contrastive}&={1\over 2} (H(\Tilde{y}_{s2p},s_{s2p})+H(\Tilde{y}_{p2s},s_{p2s})+H(\Tilde{y}_{s2s},s_{s2s})+H(\Tilde{y}_{p2p},s_{p2p}))\label{Eq10}\\
\Tilde{y}_n^{NWP}&=(1-\alpha)y_n^{NWP}+\alpha p^{NWP}_{momentum}(s_n|s_{0:n-1},P)\label{Eq11}\\
\Tilde{L}_{NWP}&=\sum^{n}_{i=1}H(\Tilde{y}_n^{NWP}, p^{NWP}(s_n|s_{0:n-1},P))\label{Eq12}
\end{align}
After the student model's parameters ${w_{model}}$ are updated for each batch, the parameters of the momentum teacher model ${w_{momentum}}$ are updated by the exponential moving average (EMA) using ${w_{model}}$ and an EMA hyperparameter $\lambda$ according to Equation \ref{Eq13}.
\begin{align}
\label{Eq13}
w_{momentum}=(1-\lambda) w_{model}+\lambda w_{momentum}
\end{align}

The overall pre-training objective is the combined loss of Contrastive, NWP, NPP, and SPM loss.
\begin{align}
\label{Eq8}
L=\Tilde{L}_{contrastive}+\Tilde{L}_{NWP}+L_{NPP}+L_{SPM}
\end{align}

\subsection{Training for downstream tasks.}
Supplementary Figure~\ref{suppl2} describes how we utilized our pre-trained model for downstream tasks. 
For PV generation and SMILES generation (Supplementary Figure ~\ref{suppl2}-(a), (b)), we don't need additional fine-tuning since their training objectives are already included in the pre-training (NWP, NPP). For the inference procedure, the model generates PV or SMILES with autoregressive sampling. Specifically, starting from the [CLS] token of the modality that we want to generate, the model predicts the first component and repeats taking the previous outputs to predict the next component until it's done or meets a sign to stop. A causal mask has to be used in the self-attention of the fusion encoder and the unimodal encoder of the generating modality to enforce the autoregressive generation.

For MoleculeNet downstream tasks and the DILI classification task that only provide SMILES data, we utilized only the SMILES encoder part of the model (Supplementary Figure ~\ref{suppl2}-(c)).
After the input molecule is encoded with the SMILES encoder, we pass the feature of the [CLS]$_S$ token through a classification/regression head to get an output. The classification/regression head consists of MLP with one hidden layer. We fine-tuned our model with the given training set and get a checkpoint with the lowest  loss on the validation set, and recorded that checkpoint's performance on the test set.

The forward reaction prediction task provides a reactant SMILES (including multiple reagent molecules) and a product SMILES. We encode these two inputs with the SMILES encoder, then feed them into the fusion encoder + prediction head. The model is trained to autoregressively generate the original product SMILES (Supplementary Figure ~\ref{suppl2}-(d)). In the inference stage, starting from the [CLS]$_S$ token, the model predicts the next token until it generates the [SEP] token. Similar to the SMILES generation, the self-attention of the fusion encoder and the reactant SMILES encoder uses a causal mask. The retro-reaction prediction task was done in the same way, but the role of the reactant and product SMILES are swapped. We fine-tuned SPMM for the forward reaction prediction task with an approach of `mixed task', meaning that the information about the major reactant is not given to the model. For both forward and retro-reaction tasks, we replaced the input reactants and products with their random non-canonical augmented SMILES\cite{smiles-aug} with a probability of 0.5.

\subsection{Data preparation.}
We obtained 20,000,000 SMILES of general molecules from PubChem\cite{kim2021pubchem} for pre-training. 
All 53 properties we used can be calculated with SMILES using the RDKit Python module\cite{rdkit}.
The dataset for the MoleculeNet downstream tasks is provided by the DeepChem\cite{deepchem} python library. We split every dataset into train/valid/test sets in a ratio of 8:1:1 using a scaffold splitter from DeepChem, which is a more harsh condition for the model than random splitting.
For the reaction prediction task, we used the USPTO-480k dataset which contains 479,035 pairs of reactants and the major product of their reaction. The retro-reaction prediction task used the USPTO-50k dataset, containing 50,037 product-reactant pairs with corresponding reaction types. Although the USPTO-50k dataset provides tags of reaction type for each reaction data, we didn't use them, following the previous retro-reaction prediction publications. 

\subsection{Implementation details.}
We employed the architecture of 6 BERT$_{base}$ encoder layers for our PV encoder and SMILES encoder, and 6 BERT$_{base}$ encoder layers with cross-attention layers for our fusion encoder. With given $Q\in\mathbb{R}^{len_q\times d_k}, K\in\mathbb{R}^{len_k\times d_k},$ and $V\in\mathbb{R}^{len_k\times d_v}$ as query, key, and value inputs, the self-attention and cross-attention layers in BERT compute the output of the scaled-dot attention according to the following formula:
\begin{align}
\label{attn}
Attention(Q,K,V)=Softmax({QK^T \over \sqrt{d_k}})V
\end{align}

We pre-trained the model until it converges using a batch size of 128 and the AdamW optimizer with a weight decay of 0.02. The learning rate is warmed up to ${1e-4}$ and decreased to ${1e-5}$ with a cosine scheduler. We used the momentum-adjusting hyperparameter $\alpha$ of 0.4. Since the pseudo-label from the momentum teacher is not useful in the early stages of the training, we linearly increased $\alpha$ from 0 to 0.4 during the first epoch. The EMA hyperparameter $\lambda$ was fixed to 0.995, and the size of the PV and SMILES queue $k$ was set to 32,768. The momentum models are not used for downstream tasks. The full description of training for downstream tasks is in Supplementary Table ~\ref{suppl3}.

 \begin{addendum}
{\color{black} \item[Correspondence] Correspondence and requests for materials should be addressed to Jong Chul Ye.~(email: jong.ye@kaist.ac.kr).}
 \item  This work was supported by the Institute of Information \& communications Technology Planning \& Evaluation (IITP) grant funded by the Korea government (MSIT)  (No.2019-0-00075, Artificial Intelligence Graduate School Program (KAIST)), National Research Foundation (NRF) of Korea grant NRF-2020R1A2B5B03001980, and by the KAIST Key Research Institute (Interdisciplinary Research Group) Project.
{
\item[Author Contributions] J.C. prepared the code, performed all experiments and analyses, collected data, and wrote the manuscript. J.C.Y. supervised the project in conception and discussion, and prepared the manuscript.}
 \item[Competing Interests] 
The authors declare that they have no competing financial interests.

\end{addendum}

\section*{Data Availability}
The pre-training dataset is publicly available in PubChem \href{https://pubchem.ncbi.nlm.nih.gov/}{https://pubchem.ncbi.nlm.nih.gov/}. ZINC15 test molecules for SMILES-to-PV generation are accessible through the \href{https://zinc15.docking.org/}{ZINC15 website}. Scaffold-split MoleculeNet datasets are available via DeepChem python module \href{https://deepchem.io/}{https://deepchem.io/}, and raw databases can be found in the MoleculeNet website \href{https://moleculenet.org/}{https://moleculenet.org/}. The DILI training and test data preparation can be found in \href{https://pubmed.ncbi.nlm.nih.gov/29788510/}{https://pubmed.ncbi.nlm.nih.gov/29788510/}. The USPTO-480k and USPTO-50k dataset is available at \href{https://github.com/wengong-jin/nips17-rexgen}{https://github.com/wengong-jin/nips17-rexgen} and \href{https://github.com/MolecularAI/Chemformer/tree/main}{https://github.com/MolecularAI/Chemformer/tree/main}. All of these data that we've used in this work are also uploaded at \href{https://github.com/jinhojsk515/SPMM/}{https://github.com/jinhojsk515/SPMM/}.

\section*{Code Availability}
The source code for SPMM, a list of 53 properties for PV, experimental codes, and datasets are available at \href{https://github.com/jinhojsk515/SPMM/}{https://github.com/jinhojsk515/SPMM/} to allow reproducing the results.

\section*{References}
\bibliographystyle{naturemag}
\bibliography{ref}

\newpage

\section*{Supplementary Materials}
\setcounter{figure}{0}
\renewcommand{\thefigure}{S\arabic{figure}}
\setcounter{table}{0}
\renewcommand{\thetable}{S\arabic{table}}

\begin{figure}[!h]
	\centering
 \centerline{\epsfig{figure=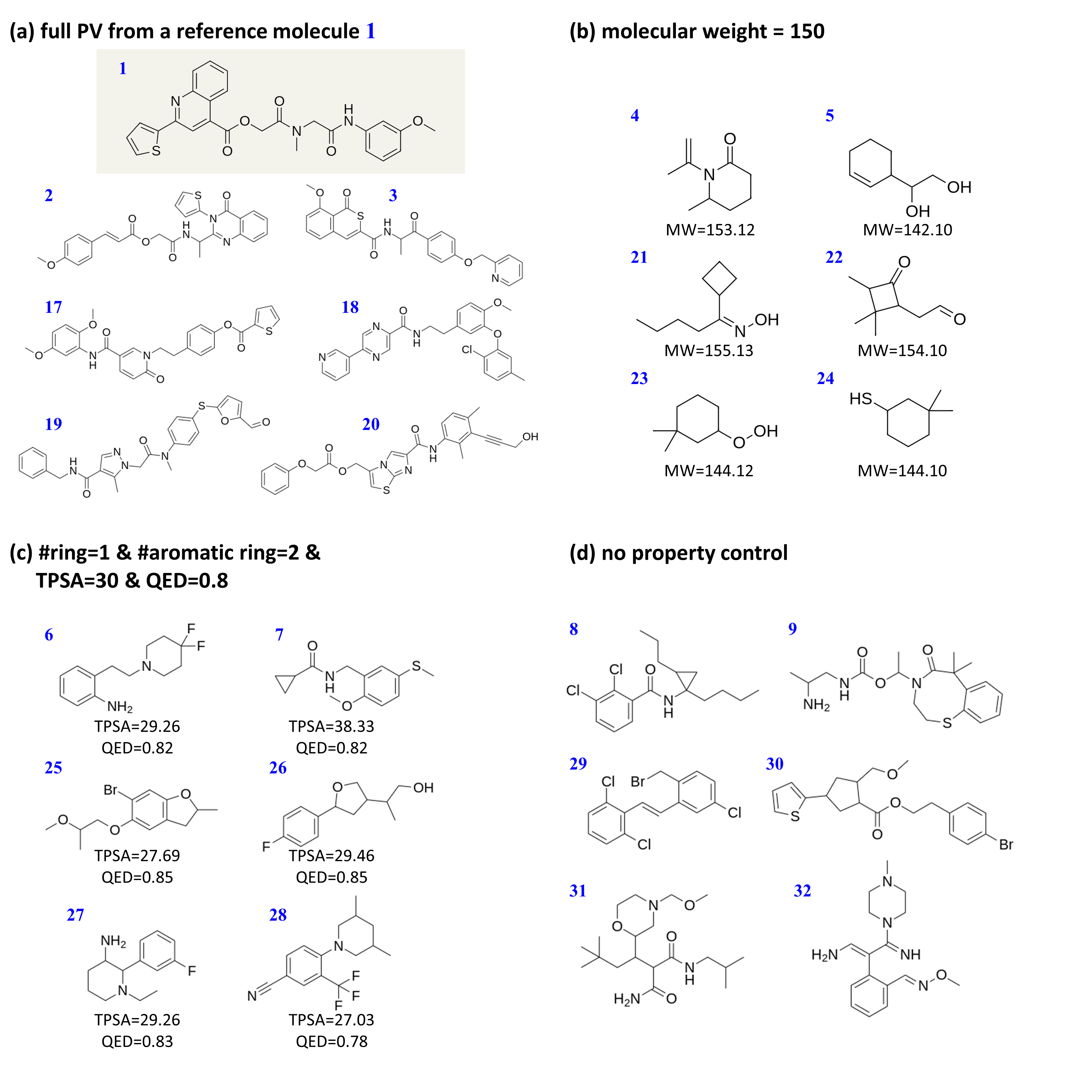, width=1\linewidth}}
	\caption{\bf\footnotesize 
The examples of the output molecules in the stochastic PV-to-SMILES generations in the Fig.~\ref{fig4}, by pre-trained SPMM with four different given PVs.}
	\label{suppl4}
\end{figure}

\begin{figure}[!h]
	\centering
 \centerline{\epsfig{figure=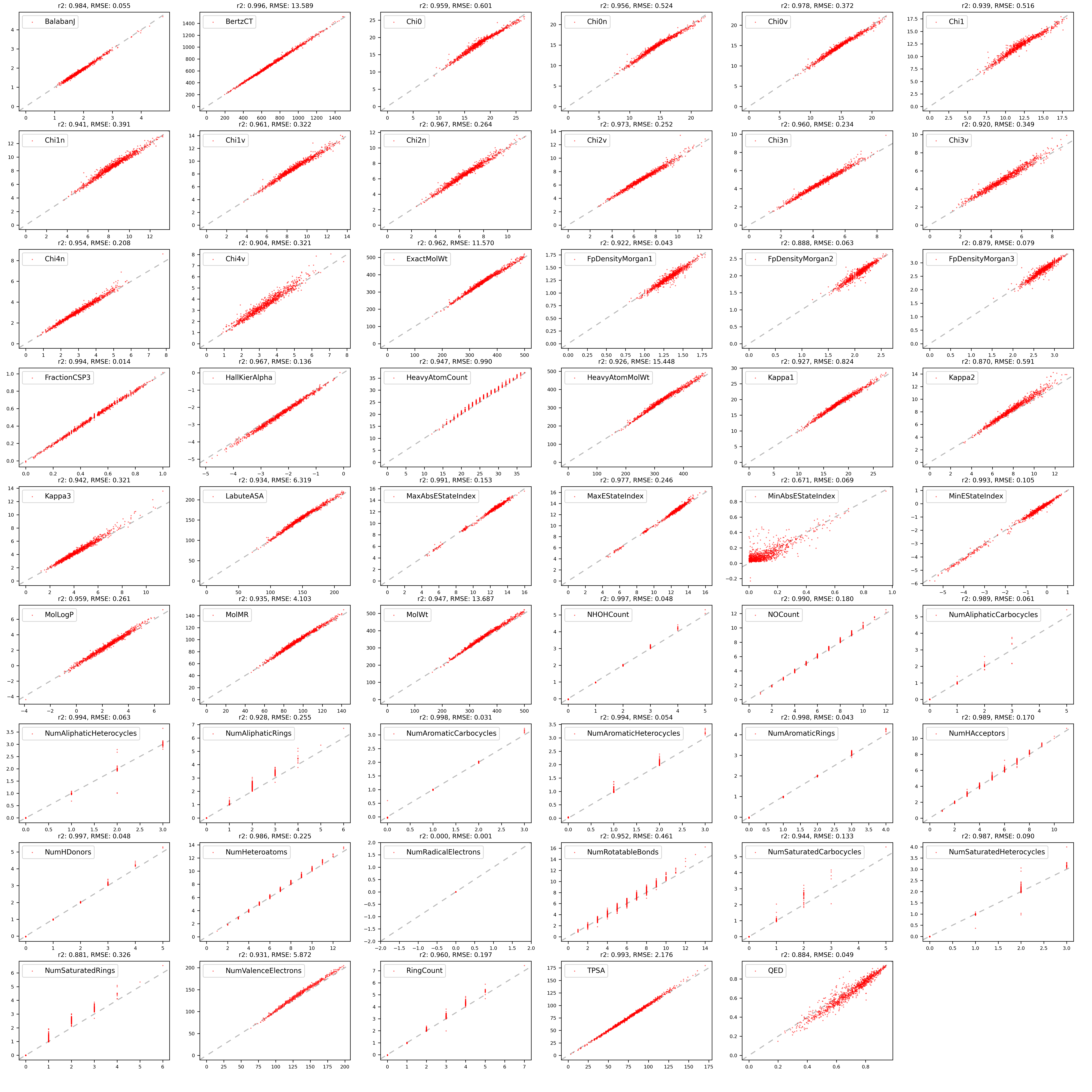, width=0.95\linewidth}}
	\caption{\bf\footnotesize 
The scatter plots of the model's generated PVs for 1,000 unseen ZINC15 molecules, against the actual property value for all 53 properties. The $r^2$ score and RMSE for each property are described at the top of each plot.
}
	\label{supp1}
\end{figure}

\begin{figure}[!h]
	\centering
 \centerline{\epsfig{figure=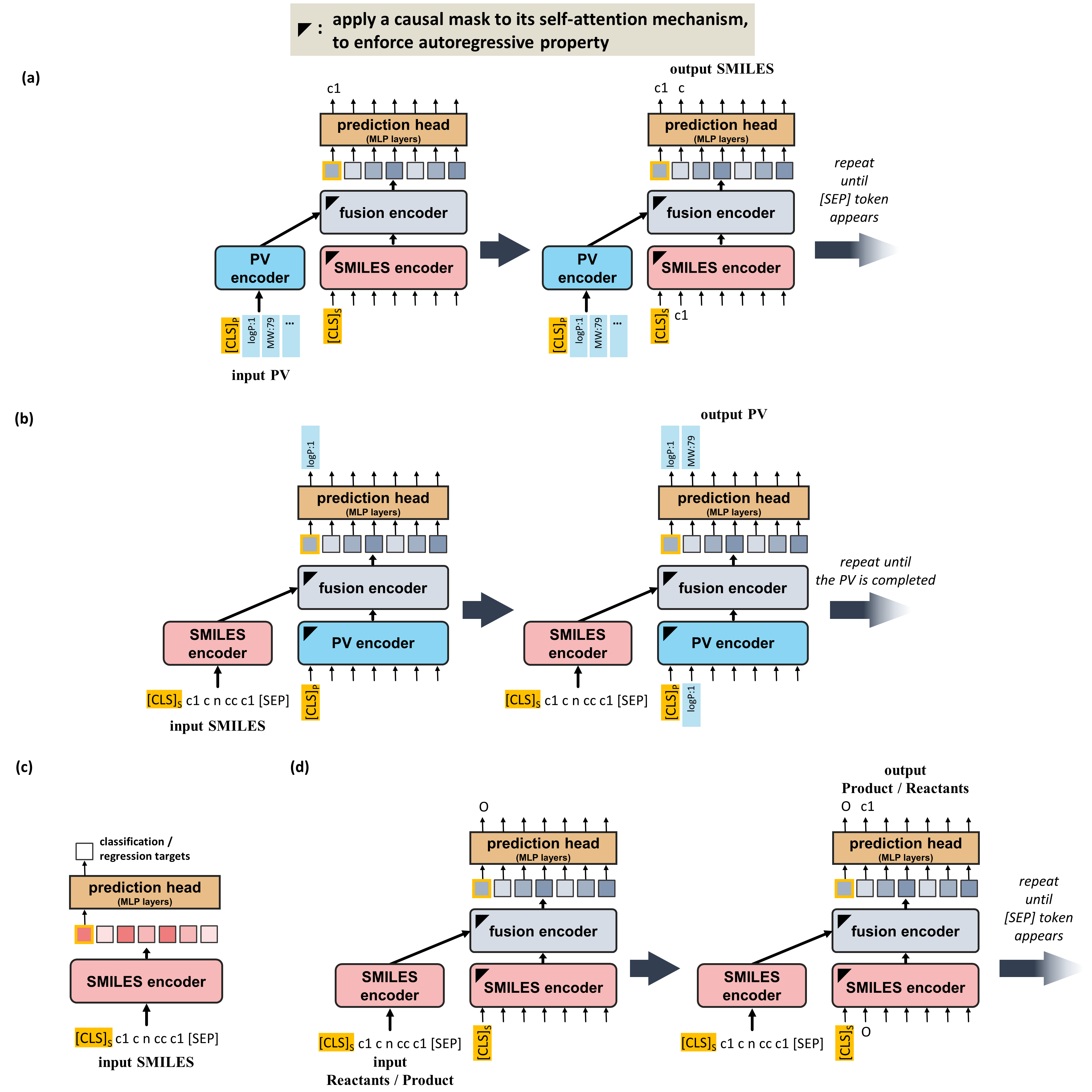, width=1.0\linewidth}}
	\caption{\bf\footnotesize 
Overview of the inference and fine-tuning of SPMM for various downstream tasks: (a) The inference process of pre-trained SPMM for molecule generation. (b) The inference process of pre-trained SPMM for PV generation. (c) The model architecture for MoleculeNet downstream tasks. The SMILES encoder of pre-trained SPMM is used as a backbone. (d) The model architecture for the reaction prediction task. We adopted the SMILES encoder and the fusion encoder of pre-trained SPMM and built a sequence-to-sequence model.
}
	\label{suppl2}
\end{figure}

\begin{table}
\centering
\resizebox{\columnwidth}{!}{%
\begin{tabular}{r|cccccccccc}
\hline 
task                 & Delaney ESOL    & LIPO            & Freesolv           & BACE(reg.)      & BACE(cls.)           & Clearance       & BBBP            & Clintox         &SIDER      &DILI\\ \hline
%optimizer               & AdamW           & AdamW           & AdamW           & AdamW            & AdamW                & AdamW           & AdamW           & AdamW           \\
%scheduler               & cosine + warmup & cosine + warmup & cosine + warmup & cosine + warmup  & cosine + warmup      & cosine + warmup & cosine + warmup & cosine + warmup \\
optimizer               & \multicolumn{10}{c}{AdamW, weight decay=0.02}              \\
scheduler               & \multicolumn{10}{c}{cosine + warmup}    \\
batch size              & 4               & 16              & 4               & 8                & 8                    & 8               & 16               & 4            &4            &8\\
learning rate(min, max) & 3e-5, 5e-6      & 3e-5, 5e-6      & 5e-5, 5e-6      & 2e-5, 2e-6       & 5e-5, 1e-5           & 2e-5, 5e-6      & 3e-5, 5e-6      & 3e-5, 5e-6    &5e-5, 5e-6   &2e-5, 5e-6 \\
epoch                   & 25              & 25              & 30              & 30               & 10                   & 10              & 10              & 15            &10           &20\\ \hline
\end{tabular}%
}
\resizebox{0.5\columnwidth}{!}{%
\begin{tabular}{r|cc}
\hline 
task                 & forward reaction prediction & retro-reaction prediction  \\ \hline
optimizer               & \multicolumn{2}{c}{AdamW, weight decay=0.02}      \\
scheduler               & \multicolumn{2}{c}{cosine + warmup}    \\
batch size              & 16          & 16    \\
learning rate(min, max) & 1e-4, 3e-6  & 1e-4, 5e-6    \\
epoch                   & 100          & 300    \\ \hline
\end{tabular}%
}
\caption{\bf\footnotesize 
Detailed training hyperparameters for fine-tuning SPMM in MoleculeNet downstream tasks, DILI classification task, forward reaction prediction task, and the reverse-reaction prediction task.
}
	\label{suppl3}
\end{table}

\end{document}

%% file: macros.tex
\long\def\comment#1{} % comment out text

\newcommand{\beq}{\begin{equation}}
\newcommand{\eeq}{\end{equation}}
\newcommand{\beqa}{\begin{eqnarray}}
\newcommand{\eeqa}{\end{eqnarray}}